\documentclass[10pt,twocolumn,letterpaper]{article}

\usepackage{cvpr}              

\usepackage{amsmath}
\usepackage{amssymb}
\usepackage{bm}
\usepackage{interval}

\usepackage{algorithm}
\usepackage{algpseudocode}
\usepackage{amsmath, amssymb}

\usepackage{graphicx}
\usepackage{caption}
\usepackage{subcaption}
\usepackage{multirow}
\usepackage{booktabs}
\usepackage{tabulary}
\usepackage{colortbl}

\usepackage{times}
\usepackage{textcmds}
\usepackage[dvipsnames]{xcolor}

\usepackage{epsfig}
\usepackage{etoolbox}

\usepackage{cuted}
\usepackage{capt-of}

\definecolor{best}{rgb}{1.0, 0.85, 0.85}      
\definecolor{second}{rgb}{1.0, 0.93, 0.75}    
\usepackage{booktabs}      
\usepackage{multirow}      
\usepackage{colortbl}      
\usepackage{xcolor}        
\usepackage{tabulary}
\usepackage{etoolbox}
\usepackage{tabularx}
\definecolor{catgray}{RGB}{240,240,240}

\newcommand{\best}[2]{%
  \ifnum#1=1
    \cellcolor{red!50}#2%
  \else\ifnum#1=2
    \cellcolor{orange!50}#2%
  \else\ifnum#1=3
    \cellcolor{yellow!50}#2%
  \else
    #2%
  \fi\fi\fi
}

\definecolor{cvprblue}{rgb}{0.21,0.49,0.74}
\usepackage[pagebackref,breaklinks,colorlinks,allcolors=cvprblue]{hyperref}

\def\paperID{} 
\def\confName{CVPR}
\def\confYear{2026}

\title{DSA: Dynamic Step Allocation for Fast Autoregressive Video Generation}

\author{%
  Thanh-Tung Le$^{1,2}$\thanks{Work done while Thanh-Tung Le is a Student Researcher at Google.} \quad
  Yunhan Zhao$^{3}$ \quad
  Menglei Chai$^{2}$ \quad
  Zhengyang Shen$^{2}$ \quad
  Zhe Cao$^{2}$ \\[0.4em]
  Danhang Tang$^{2}$ \quad 
  Xiaohui Xie$^{1}$ \quad
  Deying Kong$^{2}$
  \\[0.4em]
  $^{1}$University of California, Irvine \quad
  $^{2}$Google \quad
  $^{3}$Google DeepMind
}

\newcommand{\mname}{\texttt{DSA}\xspace}

\begin{document}
\twocolumn[{%
\renewcommand\twocolumn[1][]{#1}%
\maketitle
\vspace{-1em}

\begin{center}
    \centering
    \includegraphics[width=0.92\linewidth]{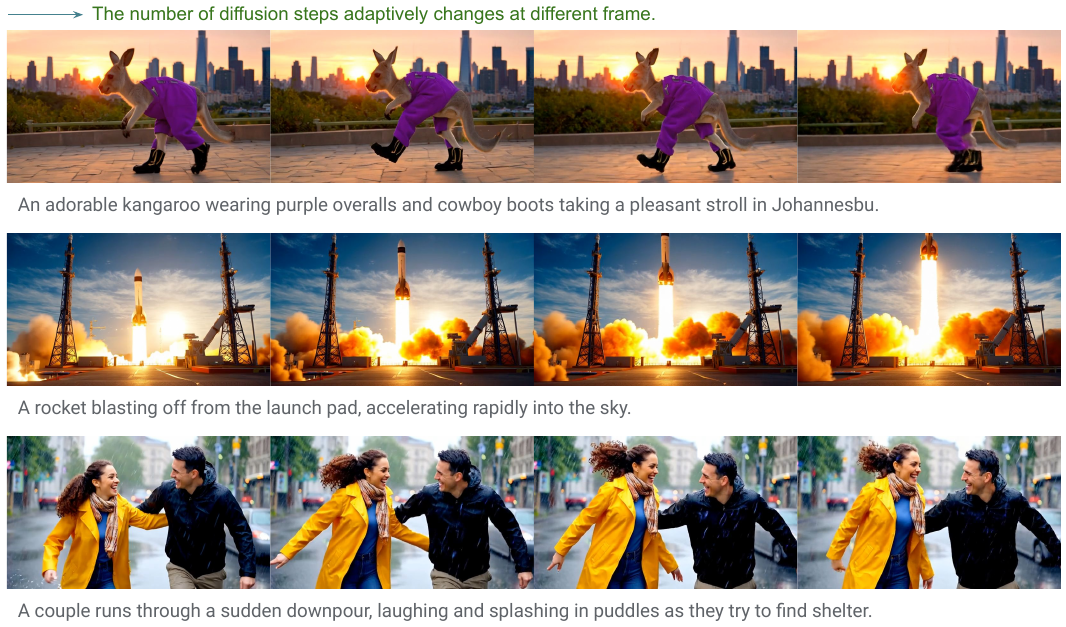}
    \captionof{figure}{Our method can dynamically adjust the number of diffusion steps when generating different frames and achieve real-time autoregressive video generation, reaching 22.63 FPS with sub-second latency on H100 GPUs.}
    \label{fig:teaser}
\end{center}
\vspace{1em}
}]


\begin{abstract}
Video diffusion transformers have achieved state-of-the-art visual quality, but their high inference cost remains a major bottleneck for real-time applications. Recent distillation frameworks produce autoregressive video diffusion models with reduced latency, yet these models still use a fixed number of denoising steps per frame, wasting computation on predictable frames and under-refining challenging ones. We present \mname, a confidence-guided adaptive computation framework for AR video diffusion. \mname introduces a lightweight confidence head, trained jointly with the generator under a distribution-matching distillation objective, to estimate per-frame denoising reliability. At inference, this confidence signal dynamically adjusts the number of diffusion steps: simple frames terminate early for speed, while complex frames receive additional refinement. Our method requires no extra video data, no heuristics, and little architectural modification. Experiments show that \mname achieves real-time autoregressive video generation, reaching 22.63 FPS with sub-second latency on H100 GPUs, while maintaining competitive or superior VBench quality compared to recent autoregressive and bidirectional video diffusion models. Our results demonstrate that confidence-guided adaptive sampling provides an effective and practical path toward interactive video generation.
\end{abstract}    
\section{Introduction}
\label{sec:intro}

Diffusion models~\cite{ho2020ddpm,song2020ddim} have become the dominant paradigm in generative modeling, surpassing VAEs~\cite{kingma2013vae,rolfe2016discretevae}, GANs~\cite{karras2019stylegan,goodfellow2020generative}, and autoregressive models~\cite{chang2022maskgit,chang2023muse} in generation quality. Their success spans diverse domains, including image~\cite{rombach2022stablediffusion,saharia2022imagen}, video~\cite{singer2022makeavideo,blattmann2023svd}, 3D~\cite{poole2022dreamfusion,liu2023zero123}, and audio~\cite{kong2020diffwave,huang2023makeanaudio} synthesis, as well as image~\cite{hertz2022p2p,avrahami2023blended} and video~\cite{qi2023fatezero,wu2023tuneavideo} editing.
Recent Diffusion Transformers (DiTs)~\cite{peebles2023dit,ma2024sit} further advance scalability and generalization beyond UNet-based diffusion models~\cite{rombach2022stablediffusion}, marking a new direction for Generative AI. However, DiTs remain computationally demanding in both memory and inference, especially with large token counts as seen in high-resolution or long video generation. This high resource requirement, speculated to underlie the limited deployment of models like Sora~\cite{brooks2024sora,liu2024sorareview}, has motivated efforts to improve efficiency, including latent diffusion~\cite{rombach2022stablediffusion}, step distillation~\cite{sauer2023adversarial,yin2024one}, caching~\cite{wimbauer2024blockcache,ma2024deepcache,habibian2024clockwork}, architecture search~\cite{zhao2023mobilediffusion,li2024snapfusion}, token reduction~\cite{bolya2023tomesd,li2024vidtome}, and region-based generation~\cite{nitzan2024lazydiff,kahatapitiya2024ocd}.

In video generation, particularly for interactive applications, not only generation speed and temporal consistency but also inference latency are critical factors~\cite{huang2025self}. While caching-based acceleration~\cite{liu2025timestep, kahatapitiya2025adaptive} methods help reduce computation in bidirectional foundation models, their reliance on expensive full-context attention makes them less suitable for real-time applications. In contrast, autoregressive video diffusion models generate frames sequentially with causal attention, enabling efficient streaming.


Recent studies~\cite{yin2025slow, huang2025self} attempt to distill bidirectional foundation video diffusion models into autoregressive variants, enabling the use of KV-caching for step-wise denoising over short temporal segments. This design facilitates efficient streaming and interactive generation. Notably, Self Forcing (SF)~\cite{huang2025self} introduces autoregressive self-rollout with KV-caching during both training and inference, effectively distilling bidirectional diffusion into few-step autoregressive diffusion while balancing computational efficiency and visual fidelity.

Despite these advances, prior AR models rely on a fixed number of denoising steps per frame, regardless of how easy or difficult the frame is to generate. However, video sequences naturally vary in temporal complexity: predictable or low-motion frames require minimal refinement, while frames containing rapid motion or structural changes benefit from additional denoising. Using a uniform sampling budget thus wastes computation on trivial frames and under-serves challenging ones.



We address this limitation with \mname, short for \textbf{D}ynamic \textbf{S}tep \textbf{A}llocation for Fast Autoregressive Video Generation,  a confidence-guided adaptive computation method for accelerating autoregressive video diffusion. A lightweight confidence network, trained jointly with the transformer during distillation, predicts chunk-wise difficulty and adjusts sampling accordingly at inference. Unlike previous caching-based approaches~\cite{kahatapitiya2025adaptive, liu2025timestep}, our method requires neither additional video data for finetuning nor handcrafted heuristics. Instead, it leverages the distribution-matching distillation framework~\cite{yin2024improved, yin2024one}, which allows joint optimization of the transformer and confidence network using text data alone. Extensive experiments demonstrate that \mname enables real-time autoregressive video generation. Distilled from both Wan-1.3B and Wan-14B models~\cite{wan2025wan}, \mname achieves on average 22.63 FPS on an H100 GPU in throughput, while retaining competitive visual quality, measured in VBench~\cite{huang2024vbench} benchmark.

Our main contributions include the following:
\begin{itemize}
    \item A confidence-guided adaptive sampling framework for autoregressive video diffusion, which dynamically adjusts the number of denoising steps per frame using a lightweight confidence head trained jointly with the diffusion transformer, requiring no extra data, no heuristics, and little architectural changes.
    \item A unified distillation approach combining Distribution-Matching Distillation with a quality-aware confidence objective, enabling efficient few-step autoregressive generation that achieves real-time performance while matching or surpassing the visual quality of state-of-the-art bidirectional and autoregressive video diffusion models.
\end{itemize}
\section{Related works}
\label{sec:related-works}

\begin{figure*}[t]
    \centering
    \includegraphics[width=0.95\textwidth]{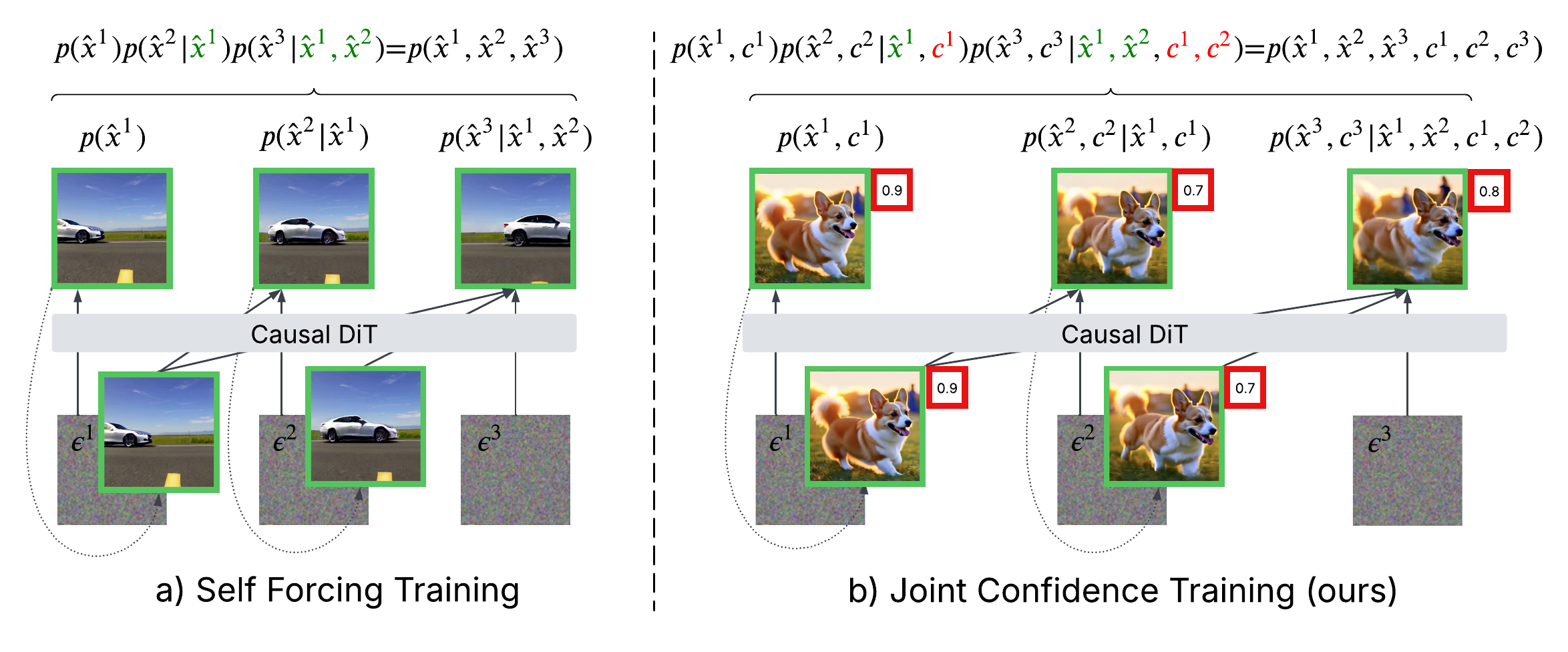}
    \caption{Training paradigms comparison between the Self Forcing~\cite{huang2025self} and our method. Left figure is taken from~\cite{huang2025self}. The Self Forcing approach performs autoregressive self-rollout during training, denoising the next frame based on previous context frames generated by itself. Our method further generates the confidence score during the roll-out and conditions the denoising on the confidence score as well.}
    \label{fig:training-paradigms}
\end{figure*}

\subsection{Autoregressive/Bidirectional Diffusion Models for Video Generation}

Early video generation methods primarily used GANs~\cite{goodfellow2020generative}, either generating full videos with convolutional models~\cite{brooks2022generating, saito2017temporal} or producing frames sequentially with recurrent architectures~\cite{denton2017unsupervised, li2022infinitenature, liu2021infinite}. GANs have also been adopted for distilling video diffusion models~\cite{lin2025diffusion, mao2025osv, wu2025snapgen}. Because GAN generators follow the same procedure during training and inference, they naturally avoid exposure bias—a principle we leverage by directly aligning the generator’s output distribution with the target data. Modern video generation has since shifted toward diffusion and autoregressive models for their scalability: diffusion models denoise all frames jointly using bidirectional attention~\cite{blattmann2023svd, blattmann2023align, gupta2024photorealistic, hacohen2024ltx}, while autoregressive models use next-token prediction and generate spatiotemporal tokens sequentially at inference~\cite{bruce2024genie, kondratyuk2023videopoet, ren2025next}.

\subsection{Distribution Matching Distillation from Bidirectional to Autoregressive Video Generation}
A major direction for accelerating video diffusion models is distillation from large multi-step foundation models into efficient few-step students~\cite{yin2024improved, yin2024one}. Yin et al.~\cite{yin2024improved, yin2024one} introduced Distribution Matching Distillation (DMD) for compressing multi-step image diffusion models, and subsequent works extended DMD to video generation~\cite{yin2025slow, huang2025self}. CausVid~\cite{yin2025slow} proposed asymmetric distillation—training a causal student using a bidirectional teacher—while Self-Forcing~\cite{huang2025self} further reduced train–test mismatch through rolling KV cache training. Our approach is closest to Self-Forcing but introduces a novel adaptive sampling strategy that significantly improves sampling speed without sacrificing generation quality.

\section{Method}
\label{sec:method}

In this section, we first review autoregressive video diffusion and the Self Forcing (SF) distillation framework in Sec~\ref{subsec:prelim}, highlighting how few-step AR diffusion with KV caching enables efficient causal generation. We then introduce our key extension, i.e., jointly modeling video frames and per-frame confidence scores, allowing the generator to assess its own denoising reliability during the rollout in Sec~\ref{subsec:training}. Next, we describe our confidence-augmented diffusion formulation, including the confidence head architecture and our quality-aware confidence loss that aligns confidence with generation quality under the DMD training objective. Finally, we present our confidence-guided inference algorithm, which adaptively adjusts the number of denoising steps per frame, yielding significant computational savings while maintaining temporal coherence and visual fidelity in Sec~\ref{subsec:inference}.
\subsection{Preliminaries}
\label{subsec:prelim}

{\bf Autoregressive (AR) video diffusion models} combine diffusion-based synthesis quality with autoregressive temporal causality. They factorize the video frame distribution $x^{1:N} = (x^1, \ldots, x^N)$ into conditional distributions $p(x^{1:N}) = \prod_{i=1}^N p(x^i|x^{<i})$, modeling each term via diffusion. Each frame is generated by denoising Gaussian noise conditioned on previous frames, capturing long-range temporal dependencies while maintaining visual quality. Generation can occur frame-by-frame or in temporal chunks to balance context and efficiency~\cite{yin2025slow, teng2025magi}. Autoregressive video diffusion models use frame-wise or chunk-wise denoising with Teacher Forcing (TF) or Diffusion Forcing (DF)~\cite{song2025history} paradigms. Specifically, each frame $x^i$ is corrupted by the forward process $q_{t_i \mid 0}(x^i_{t_i} \mid x^i_0)$ such that $x^i_{t_i} = \Psi(x^i, \epsilon^i, t_i) = \alpha_{t_i} x^i + \sigma_{t_i} \epsilon^i$, where $\alpha_{t_i}, \sigma_{t_i}$ are pre-defined noise schedules within a finite time 
horizon $t_i \in [0, 1000]$ and $\epsilon^i \sim \mathcal{N}(0, I)$. TF learns to predict clean frames from ground-truth context, while DF uses noise-corrupted context to simulate inference. Formally, TF models the conditional distribution for the $i$th frame at noise level $t_j$ as $p(x^i_{t_j} | x_0^{<i})$, where all conditional history frames are the ground-truth clean frames from the training data. On the other hand, DF models $t_j$ as $p(x^i_{t_j} | x^{<i}_{t \geq 0})$, where the history frames are the ground-truth frames corrupted with independent noise levels. Since training relies on ground-truth histories while inference relies on the model’s own predictions, a train–test gap known as exposure bias arises~\cite{schmidt2019generalization, yin2025slow}.

Self Forcing~\cite{huang2025self}, an autoregressive distillation method, directly addresses this mismatch through autoregressive self-rollout during training: each frame is denoised based on previously self-generated outputs, mirroring inference. For computational tractability, SF uses few-step diffusion with gradient truncation, backpropagating only through the final denoising step per frame. Randomly sampling denoising steps per iteration distributes supervision across intermediate steps, improving stability and efficiency. The method also incorporates KV-caching during training, previously used only at inference, to maintain causal memory across frames without specialized attention masks. Crucially, SF enables holistic, video-level supervision by aligning generated video distributions $p_\theta(x^{1:N})$ with real data $p_{\text{data}}(x^{1:N})$ through distribution-matching objectives. Rather than optimizing per-frame denoising losses, it applies divergence measures between entire video distributions, such as Distribution Matching Distillation (DMD)~\cite{yin2024one}, Score Identity Distillation (SiD)~\cite{zhou2024score}, and GAN~\cite{goodfellow2020generative}. This holistic formulation enforces temporal consistency and mitigates exposure bias by training under the model's own generative distribution. Our method is built upon SF with DMD objective. We aim to further improve sampling speed while maintaining high visual quality. To achieve this, we propose an approach that efficiently performs adaptive computation during the generation process. 

\subsection{Joint AR Modeling of Frames and Confidence}
\label{subsec:training}


To enable adaptive computation during video generation, we extend the SF training framework by introducing per-frame confidence signals. Our core motivation is to train the model to assess its own generation quality in real-time, thus allowing the diffusion process to dynamically adjust the number of denoising steps based on convergence during inference. We model the joint generation of video frames and their corresponding reliability scores within the same AR rollout. Specifically, as illustrated in Fig.~\ref{fig:training-paradigms}, for a sequence of $N$ frames, we factorize the joint probability as
\begin{equation}
    p_\theta(x^{1:N}, c^{1:N}) = \prod_{i=1}^{N} p_\theta(x^i, c^i \mid x^{<i}, c^{<i}),
\end{equation}
where $x^i$ represents the $i$-th frame and $c^i \in \mathbb{R}$ denotes a scalar confidence score, where the history confidence $c^{<i}$ are implicitly embedded via KV-cache of previous frames. This factorization parallels SF's chain rule for frames but augments it with confidence estimates, maintaining the established causal conditioning structure and KV-cache semantics while providing the necessary signals for adaptive step control.

\paragraph{Few-Step AR Diffusion with Confidence Head.} 
We define the autoregressive model distribution through a sequence of denoising transitions applied to an initial Gaussian latent. Specifically, the conditional distribution for frame $x^i$ is implicitly parameterized as:
\begin{equation}
    p_\theta(x^i_0 \mid x^{<i}, c^{<i})
    \;\triangleq\;
    \big(f_{\theta,t_1} \circ f_{\theta,t_2} \circ \dots \circ f_{\theta,t_T}\big)(x^i_{t_T}),
\end{equation}
where $f_{\theta,t_j}(x^i_{t_j}) = \Psi\!\big(G_\theta(x^i_{t_j}, t_j, x_0^{<i}, c^{<i}),\, \epsilon_{t_{j-1}},\, t_{j-1}\big)$, $x^i_{t_T} \sim \mathcal{N}(0, I)$, and the generator outputs both the denoised prediction and its confidence: 
\begin{equation}
    (\hat{x}^i_{0,t_j},\, c^i_{t_j})
    =
    G_\theta(x^i_{t_j},\, t_j,\, x_0^{<i},\, c^{<i}).
\end{equation}

\begin{figure}[htbp]
    \centering
    \includegraphics[width=0.47\textwidth]{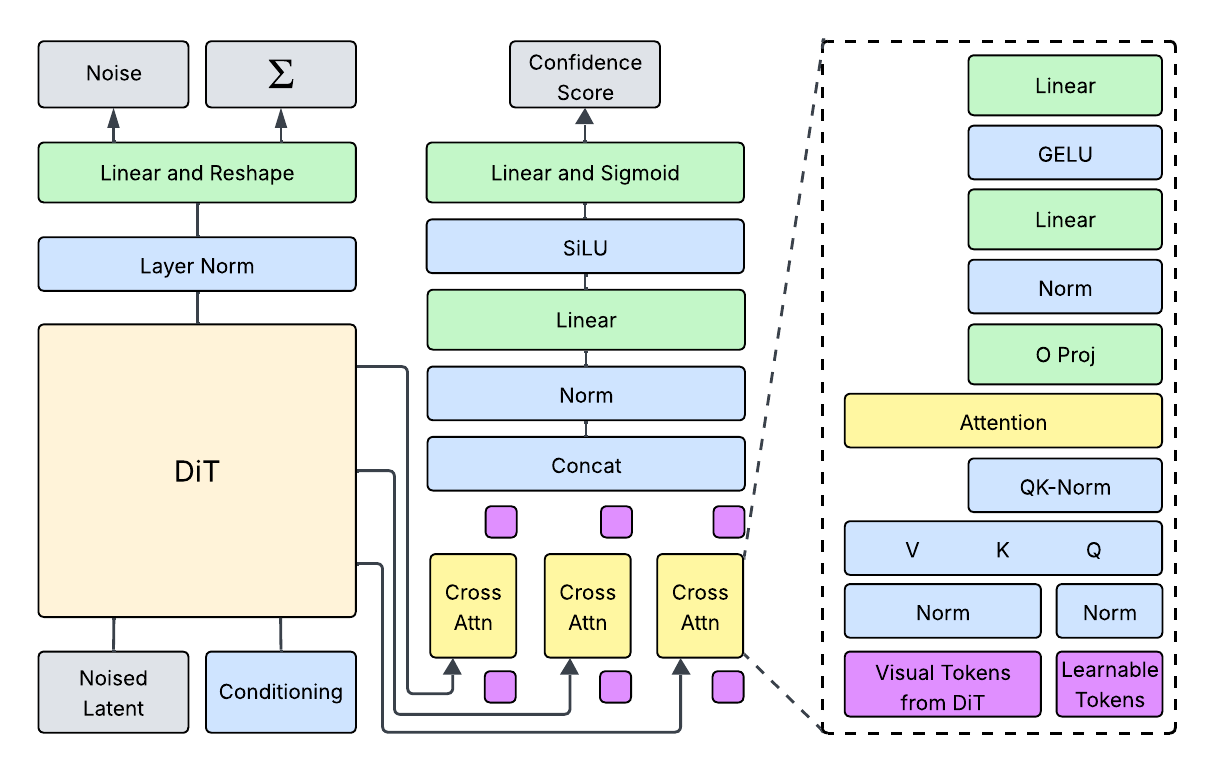}
    \caption{Model architecture for confidence score. Cross-attention is performed between learnable tokens and the visual tokens from the transformer backbone. After that, the tokens are mapped to a scalar confidence score.}
    \label{fig:confidence-head}
\end{figure}

To predict whether the diffusion process has converged sufficiently so that it can terminate early, we modify the diffusion transformer to produce an additional confidence logit, as shown in Fig.~\ref{fig:confidence-head}.  Cross-attention is applied at three specific stages of the generator’s transformer backbone. Specifically, we extract and process features from DiT blocks 13, 21 and 29. Each block utilizes a single learnable token as the query to cross-attend to all visual tokens, producing a concise token output. These outputs are channel-concatenated, normalized, and projected to yield a single scalar logit, which is finally converted to a confidence score via the sigmoid function. Crucially, unlike~\cite{lin2025diffusion} that relies on a separate critic network, we autoregressively extract logits directly from the generator's backbone. This design enables the simultaneous output of the denoised frame $x^{i}_0$ and its confidence score $c^{i}$ with negligible computational overhead, facilitating efficient, dynamic sampling at inference time.


During training, both $x^{i}_0$ and $c^{i}$ are employed as the autoregressive context, and their KV embeddings are stored in the cache for subsequent frames. This yields a causal conditioning structure shared between training and inference.

\paragraph{Quality-Aware Confidence Loss.}
To ensure that the confidence score accurately reflects the generation quality, thereby acting as a reliable proxy for step adaptation, we propose a quality-aware confidence loss. 
With a slight abuse of notation, let $\hat{\bf x}^{G} := [\hat x^1_0, \hat x^2_0,..., \hat x^n_0]$ be the set of predicted frames and $ {\bf c}:=[c^0, c^1, ..., c^n]$ be the corresponding confidence scores predicted by the generator. We define the target distribution sequence via a teacher model as
$\hat{\bf x}^{R} := \big[ R_{\phi}(x^i_t, t) \big]_{i=0}^n$,
where $x^i_t = \Psi(x^i_0, \epsilon_t, t)$ is the noisy latent for the $i$-th frame, and $R_\phi(\cdot)$ is the score function estimated by the pre-trained teacher diffusion model, referred to as the real score network~\cite{yin2024improved}.
The training loss for the confidence head is then formulated as
\begin{equation}\label{eq:conf-loss}\mathcal{L} = {\bf c} \odot \text{sg}[d(\hat{\bf x}^{G}, \hat{\bf x}^{R})] - \lambda \log {\bf c},\end{equation}
where $\operatorname{sg}[\cdot]$ denotes the stop-gradient operator, $\lambda$ is a regularization coefficient~\cite{wan2018confnet}, and $d(\cdot)$ is the L2 distance metric between the generated and teacher-predicted samples.
The loss is averaged over the sequence of predicted frames.

The final training objective incorporates both the above confidence loss and the standard DMD loss.
Intuitively, whereas the DMD loss~\cite{yin2024improved,yin2024one} forces the generator to match the teacher by minimizing the residual between $\hat{\bf x}^{G}$ and $\hat{\bf x}^{R}$, our confidence loss leverages this residual to calibrate the confidence score $\bf c$, ensuring that it accurately reflects the generation divergence. When the student model generates high quality frames, i.e., the $\hat{\bf x}^{G}$ closely matches $\hat{\bf x}^{R}$, the model is incentivized to produce high confidence score to reduce the regularization penalty.
In contrast, significant deviations between $\hat{\bf x}^{G}$ and $\hat{\bf x}^{R}$ would result in lower predicted confidence. This joint optimization ensures that the generator learns self-calibrated uncertainty, empowering the sampler to adaptively reduce diffusion steps when high confidence is achieved during inference.


\subsection{Autoregressive Diffusion Inference with Confidence-Guided Dynamic Diffusion Steps}
\label{subsec:inference}

    

\begin{algorithm}[t]
\caption{Autoregressive Diffusion Inference with Confidence-Guided Dynamic Diffusion Steps}
\small
\begin{algorithmic}[1]
  \Require KV cache of size $L$ frames
  \Require Denoise timesteps $\{t_1, \dots, t_T\}$
  \Require Number of generated frames $M$
  \Require AR diffusion model $G_\theta$ (returns both prediction and confidence, and KV embeddings via $G_\theta^{\mathrm{KV}}$)
  \State Initialize model output $\mathcal{X}_\theta \gets [\,]$
  \State Initialize KV cache $\mathcal{K} \gets [\,]$
  \State Initialize previous confidence $\mathrm{prev\_c} \gets 1.0$

  \For{$i = 1$ \textbf{to} $M$}
    \State Initialize $x^i_{t_T} \sim \mathcal{N}(0, I)$
    \State $\hat{x}^i_{0},\, c^i \gets G_\theta(x^i_{t_T}; t_T, \mathcal{K})$

    \If{$\text{sigmoid}(c^i) < \mathrm{prev\_c}$}
      \For{$j = T-1$ \textbf{downto} $1$}
        \State Sample $\epsilon \sim \mathcal{N}(0, I)$
        \State $x^i_{t_{j-1}} \gets \Psi(\hat{x}^i_{0}, \epsilon, t_{j-1})$
        \State $\hat{x}^i_{0},\, c^i \gets G_\theta(x^i_{t_{j-1}}; t_{j-1}, \mathcal{K})$
      \EndFor
    \EndIf

    \State $\mathrm{prev\_c} \gets \text{sigmoid}(c^i)$
    \State $\mathbf{k}^i \gets G_\theta^{\mathrm{KV}}(\hat{x}^i_{0}; 0, \mathcal{K})$

    \If{$|\mathcal{K}| = L$}
      \State $\mathcal{K}.\text{pop}(0)$ \Comment{Evict oldest entry}
    \EndIf
    \State $\mathcal{K}.\text{append}(\mathbf{k}^i)$

    \State $\mathcal{X}_\theta.\text{append}(\hat{x}^i_{0})$
  \EndFor

  \State \Return $\mathcal{X}_\theta$
\end{algorithmic}
\label{alg:inference}
\end{algorithm}

During inference, our model generates frames autoregressively, adhering to the same causal rollout used in training.
For each frame $i$, the generator $G_\theta$ predicts both the denoised frame $\hat{x}^i_0$ and its confidence logit $c^i$, initialized from Gaussian noise $x^i_{t_T} \sim \mathcal{N}(0,I)$.
The confidence score acts as an internal reliability estimate that dynamically determines how many denoising steps to perform.

Specifically, the model first produces $(\hat{x}^i_0, c^i)$ at timestep $t_T$. Given the confidence, we could simply use a fixed threshold to decide the number of steps the model should run for each frame.
However, this is suboptimal since we observe that the threshold varies by frames and by videos. 
Therefore, we opt for a method that can dynamically decide how many steps we should run per frame of each video. We observe that the quality of subsequent frames largely depends on the quality of previous frames since model is run in autoregressive manner. Furthermore, cleaner frames have higher confidence score than the artifact frames, consistent with our training goal mentioned in Section~\ref{subsec:training}. Therefore, if the predicted confidence $c^i$ is lower than that of the previous frame, the model continues denoising with progressively smaller timesteps ${t_{T-1},\ldots,t_1}$ until the output stabilizes. High-confidence frames thus require fewer refinement steps, while low-confidence ones receive additional denoising for improved fidelity. This adaptive scheme enables efficient, quality-aware video synthesis without sacrificing temporal coherence. Following~\cite{huang2025self}, a rolling KV cache is maintained to propagate temporal context across frames.
After generation, the resulting sequence $\mathcal{X}_\theta = [\hat{x}^1_0,\ldots,\hat{x}^M_0]$ forms the final video.

Algorithm~\ref{alg:inference} summarizes the procedure. By leveraging confidence for dynamic sampling, our inference achieves adaptive compute allocation, significantly reducing average sampling cost while maintaining high-quality results.

\begin{table*}[!ht]
  \small
  \caption{\textbf{Comparison with relevant baselines.} Our models, distilled from Wan-1.3B and Wan-14B teachers, achieve the highest throughput and competitive or superior VBench scores while maintaining sub-second latency. Compared to foundation AR diffusion models, e.g. SkyReels-V2, MAGI-1 ~\cite{teng2025magi, chen2025skyreelsv2infinitelengthfilmgenerative}, our approach is over an order of magnitude faster. Relative to distillation-based baselines, e.g. CausVid, Self Forcing~\cite{yin2025slow, huang2025self}, \mname delivers similar or improved visual quality with substantially higher FPS. Red cells denote best and yellow cells denote second-best.}
  \vspace{0.5em}
  \label{tab:main}
  \centering

  \begin{tabularx}{\textwidth}{l c c c *{6}{X}}
      \toprule
      \multirow{2}{*}{Model} & 
      \multirow{2}{*}{\#Params} & 
      \multirow{2}{*}{Throughput (FPS)} & 
      \multirow{2}{*}{Latency (s)} &
      \multicolumn{6}{c}{Evaluation scores $\uparrow$} \\
      \cmidrule(lr){5-10}
       & & & &
       Aesthetic Quality & Spatial Rel. & Temporal Style  & Quality Score & Semantic Score & Total Score \\
      \midrule

      \rowcolor{catgray}
      \multicolumn{10}{l}{\textit{Autoregressive Diffusion Models} $^{\dagger}$}\\
      SkyReels-V2~\cite{chen2025skyreelsv2infinitelengthfilmgenerative}  & 1.3B & 0.49 & 112 & \textendash  & \textendash & \textendash & 84.70 & 74.53 & 82.67 \\
      MAGI-1~\cite{teng2025magi}                  & 4.5B & 0.19 & 282 & \textendash & \textendash & \textendash & 82.04 & 67.74 & 79.18 \\
      \midrule

      \rowcolor{catgray}
      \multicolumn{10}{l}{\textit{Autoregressive Distillation Models}}\\
      CausVid~\cite{yin2025slow}  
        & 1.3B 
        & 15.30 
        & \cellcolor{best}0.90 
        & 65.01 
        & 71.16 
        & \cellcolor{second}24.16 
        & 83.97 
        & 79.08 
        & 82.99 \\

      Self Forcing~\cite{huang2025self}    
        & 1.3B 
        & 15.30 
        & \cellcolor{best}0.90 
        & \cellcolor{second}65.84 
        & \cellcolor{second}80.12
        & \cellcolor{best}24.33
        & \cellcolor{best}84.89
        & \cellcolor{second}80.20
        & \cellcolor{best}83.95 \\
      \midrule
      
      Ours (1.3B-Teacher)                  
        & 1.3B 
        & \cellcolor{best}22.63 
        & \cellcolor{second}0.91 
        & 64.72 
        & 76.87 
        & 24.13 
        & \cellcolor{second}84.75 
        & 79.67 
        & \cellcolor{second}83.73 \\

      Ours (14B-Teacher)                   
        & 1.3B 
        & \cellcolor{second}22.01 
        & \cellcolor{second}0.91 
        & \cellcolor{best}67.02 
        & \cellcolor{best}84.82 
        & 24.12 
        & 84.50 
        & \cellcolor{best}81.73 
        & \cellcolor{best}83.95 \\
      \bottomrule
  \end{tabularx}

  \vspace{2pt}
  {\footnotesize\raggedright$^{\dagger}$ Results taken from \cite{huang2025self}.\par}
  \vspace{-1em}
\end{table*}

\section{Experiments}
\label{sec:experiments}
In this section, we first describe our experimental setup, including model initialization, training procedure, dataset preparation, and optimization details. We then outline the evaluation metrics used to assess real-time performance and visual quality, focusing on throughput, latency, and VBench scores. Next, we compare our distilled models against both foundation autoregressive diffusion models and prior distillation-based approaches, demonstrating substantial improvements in speed and competitive or superior quality across multiple dimensions. Finally, we conduct ablation studies to analyze the effect of adaptive sampling, showing that confidence-guided step selection consistently outperforms fixed-step baselines by achieving higher quality at significantly lower computational cost.

\subsection{Experimental Details}
\label{subsec:exp-details}
We implement our method using Wan2.1-T2V-1.3B~\cite{wan2025wan} as the base model. Following prior work~\cite{yin2025slow, huang2025self}, we initialize with causal attention masking using 16K ODE solution pairs sampled from the base model. Text prompts for both initialization and training are drawn from a filtered and LLM-augmented version of VidProM~\cite{wang2024vidprom}. Different from Self-Forcing~\cite{huang2025self}, our approach employs dynamic-step diffusion with chunk-wise autoregressive generation, producing 3 latent frames per chunk. We perform optimization using the DMD objective~\cite{yin2024improved, yin2024one} combined with our confidence loss (Equation~\ref{eq:conf-loss}), with regularization coefficient $\lambda=0.2$. Training runs for 1000 steps with batch size 128 on NVIDIA H100 GPUs. We use AdamW~\cite{loshchilov2017decoupled} optimization with learning rates of $2.0 \times 10^{-6}$ for the generator and $4 \times 10^{-7}$ for the critic, updating the generator every 5 critic steps.

\subsection{Evaluation Metrics}
\label{subsec:metric}
We evaluate our method using VBench~\cite{huang2024vbench} to assess visual quality and semantic alignment. Following~\cite{huang2025self}, our evaluation measures both throughput and first-frame latency to comprehensively characterize real-time performance, with all benchmarks performed on NVIDIA H100 GPU.

\subsection{Comparison with Existing Baselines}
\label{subsec:comparison}

\begin{figure*}[htbp]
    \centering
    \includegraphics[width=0.83\textwidth]{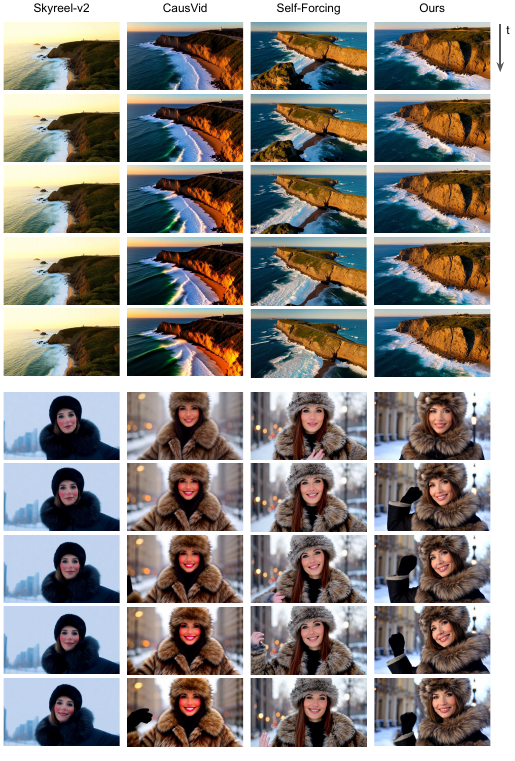}
    \caption{\textbf{Qualitative results comparison between our method and the SOTA methods.} CausVid shows noticeable error accumulation, leading to oversaturated and drifting frames over time. Our method produces sharper and more stable visuals than SkyReels-V2 while being \textbf{46x} faster in throughput and \textbf{120x} faster in latency, and it generates more realistic and temporally consistent results than Self-Forcing while running \textbf{1.5x} faster.}
    \label{fig:qualitative-results}
\end{figure*}
We evaluate our distilled models—derived from Wan-1.3B and Wan-14B—against leading autoregressive video diffusion approaches. SkyReels-V2~\cite{chen2025skyreelsv2infinitelengthfilmgenerative} and MAGI-1~\cite{teng2025magi} represent foundation AR video diffusion models trained with Diffusion Forcing, while CausVid~\cite{yin2025slow} and Self-Forcing~\cite{huang2025self} are distillation-based methods initialized from Wan-1.3B. As shown in Table~\ref{tab:main}, our models consistently match or surpass both foundation AR models and prior AR distillation methods in terms of quality and efficiency. The version distilled from Wan-1.3B achieves the highest throughput (22.63 FPS), while the Wan-14B–distilled model delivers the strongest overall accuracy, leading in Aesthetic Quality, Spatial Relationship, and Semantic Score. In contrast, foundation AR models like SkyReels-V2 and MAGI-1 run substantially slower with much higher latency. Among distillation-based AR approaches, while Self-Forcing attains strong visual fidelity, our models retain comparable or better quality while offering significantly higher speed. 

Qualitatively (Figure~\ref{fig:qualitative-results}), CausVid exhibits error accumulation, resulting in oversaturation over time. Our method produces slightly sharper visuals than SkyReels-V2 while being \textbf{46x} faster in throughput and \textbf{120x} faster in latency, and generates more realistic results than Self-Forcing while being \textbf{1.5x} faster.

\begin{figure*}[htbp]
    \centering
    \includegraphics[width=0.85\textwidth]{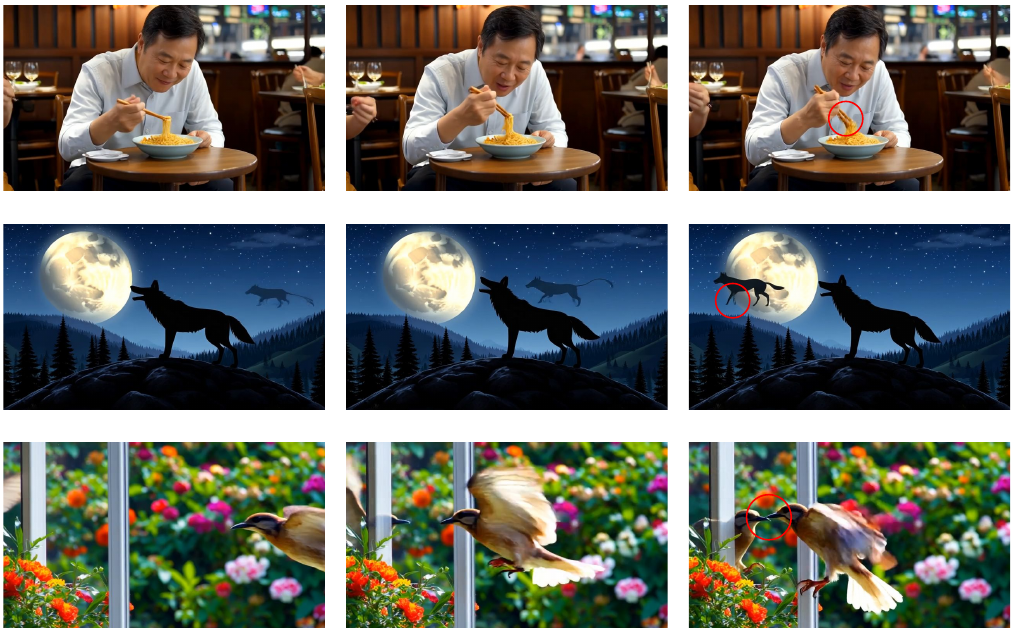}
    \caption{Illustration of some failure cases.}
    \label{fig:failure-cases}
\end{figure*}

Some failure cases are presented in Figure~\ref{fig:failure-cases}. Complex physical interactions, such as those between noodles and chopsticks, remain challenging. Similarly, accurately rendering the reflection of a bird’s beak requires sophisticated physical priors and remains difficult for the current model.

\subsection{Ablation Studies}
\label{subsec:ablation}
\begin{table}[h]
  \small
  \caption{\textbf{Ablation Study on Number of Fixed Denoising Steps.} While fixed-step methods are constrained by the speed-quality trade-off, our confidence-guided adaptive sampling identifies a superior operating point. It achieves a high VBench Total Score of 83.95 while maintaining high throughput (22.01 FPS), outperforming both low-step and high-step fixed baselines.}
  \vspace{0.5em}
  \label{tab:ablation}
  \centering

  \begin{tabularx}{0.5\textwidth}{l *{3}{>{\centering\arraybackslash}X}}
      \toprule
      Method & NFE $\downarrow$ & FPS $\uparrow$ & Total Score$\uparrow$ \\
      \midrule
      
      1-step              & 14 & 27.75 & 79.27 \\
      2-step              & 21 & 22.97 & 82.77 \\
      3-step              & 28 & 19.63 & 83.66 \\
      4-step              & 35 & 16.33 & 83.94 \\
      Ours (Adaptive)     & 23 & 22.01 & 83.95 \\
      
      \bottomrule
  \end{tabularx}

  \vspace{-1em}
\end{table}

We conduct an ablation study on how our adaptive method performs compared to a fixed number of denoising steps, as in conventional methods. Table~\ref{tab:ablation} compares fixed-step sampling with our adaptive strategy using the number of function evaluations (NFE), throughput (FPS), and VBench Total Score. Fixed-step methods exhibit the standard speed–quality trade-off: reducing the number of steps improves throughput but noticeably harms quality, while increasing the steps recovers quality at the cost of substantially slower inference. Our adaptive method avoids this compromise by dynamically adjusting the number of sampling steps based on model confidence. As a result, it achieves the highest overall quality (83.95) while maintaining significantly higher throughput (22.01 FPS) than the multi-step baselines. This demonstrates that our adaptive sampling identifies an effective sweet spot, preserving high fidelity without incurring the computational cost of fixed high-step sampling.
\section{Discussion and Conclusion}
\label{sec:discussion}

\subsection{Limitation}
\label{subsec:limitation}
While \mname significantly improves efficiency in throughput, its performance degrades on long-horizon video generation. Similar to prior autoregressive diffusion models, temporal drift and accumulated errors become more pronounced as the sequence extends, leading to reduced fidelity and semantic consistency. Addressing this limitation is an important direction for future work. Recent advances in long-context modeling~\cite{gu2024mamba} offer promising avenues for extending our adaptive framework to longer videos. Integrating these techniques with confidence-guided sampling may further enhance stability and enable high-quality generation over extended time spans.

\subsection{Conclusion}
\label{subsec:conclusion}
We presented \mname, a confidence-guided adaptive sampling framework for autoregressive video diffusion. By predicting per-frame denoising reliability, our method dynamically allocates computation and achieves real-time performance while maintaining competitive visual quality. Experiments show substantial speedups over recent AR and bidirectional diffusion models, with up to 22.63 FPS and sub-second latency on H100 GPUs. Although long-horizon generation remains challenging, our results indicate that adaptive computation is a promising direction for efficient video diffusion.

\newpage
{
    \small
    \bibliographystyle{ieeenat_fullname}
    \bibliography{main}
}

\clearpage
\setcounter{page}{1}
\maketitlesupplementary

\section{Implementation details}
Our implementation builds on open-source Wan2.1~\cite{wan2025wan} and Self-Forcing~\cite{huang2025self} frameworks, using FlashAttention-3~\cite{dao2022flashattention} for \mname. We follow the Wan2.1~\cite{wan2025wan} flow-matching parameterization with a shifted time schedule and a 4-step uniform inference schedule, following ~\cite{huang2025self}. Text prompts are curated from the VidProM~\cite{wang2024vidprom} dataset, filtered for quality and safety, and expanded using Qwen2.5-7B-Instruct~\cite{yang2024qwen25}; all VBench evaluations also employ rewritten prompts for consistency with Wan2.1. Training is performed on H100 GPUs (80GB), typically with per-GPU batch size 1 and gradient accumulation as needed to reach batch size of $128$. We initialized the real score network and critic network using pretrained weight of the base model. We train \mname using the combination of DMD loss and confidence loss given in Section~\ref{sec:method}. For DMD, the gradient of reverse Kullback-Leibler divergence is given by:

\begin{align} 
& \nabla_\theta \mathbb E_{t}[D_\text{KL}(p_{\theta, t} \| p_{\text{data}, t})] = \label{eqn:DMD-gradient} \\
& -\mathbb E_{t, \hat x_t \sim q_{t|0}(\hat x_t | \hat x), \hat x \sim p_\theta(\hat x)} \left[ (s_{\text{real}}(\hat x_t, t) - s_{\text{fake}}(\hat x_t, t))\frac{\partial \hat x}{\partial \theta}\right], \notag 
\end{align}

where $s_\text{real}(\cdot, t)$ is the score function for $p_{\text{data}, t}$, approximated by a pretrained diffusion model $f_\phi(\cdot, t)$, also referred to as the real score network, and $s_{\text{fake}}(\cdot, t)$ is the score function for $p_{\theta, t}$ and is learned through a critic network $f_\psi(\cdot, t)$ via the standard diffusion loss. The gradient in Eqn.~\eqref{eqn:DMD-gradient} is equivalent to the following loss function: 
\begin{align} 
\mathcal L_{\text{DMD}}(\theta) = \mathbb E_{t, \hat x_t, \hat x}\left[ \frac{1}{2}\left\| \hat x - \mathrm{sg}\left[\hat x - (f_\psi(\hat x_t, t) - f_\phi(\hat x_t, t))\right]\right\|^2\right], 
\end{align}
where $\mathrm{sg}[\cdot]$ denotes the stop gradient operator.

Regarding confidence loss~\ref{eq:conf-loss}, we choose a regularization coefficient $\lambda = 0.02$.

\section{Additional results}
\begin{figure}[htbp]
    \centering
    \includegraphics[width=0.45\textwidth]{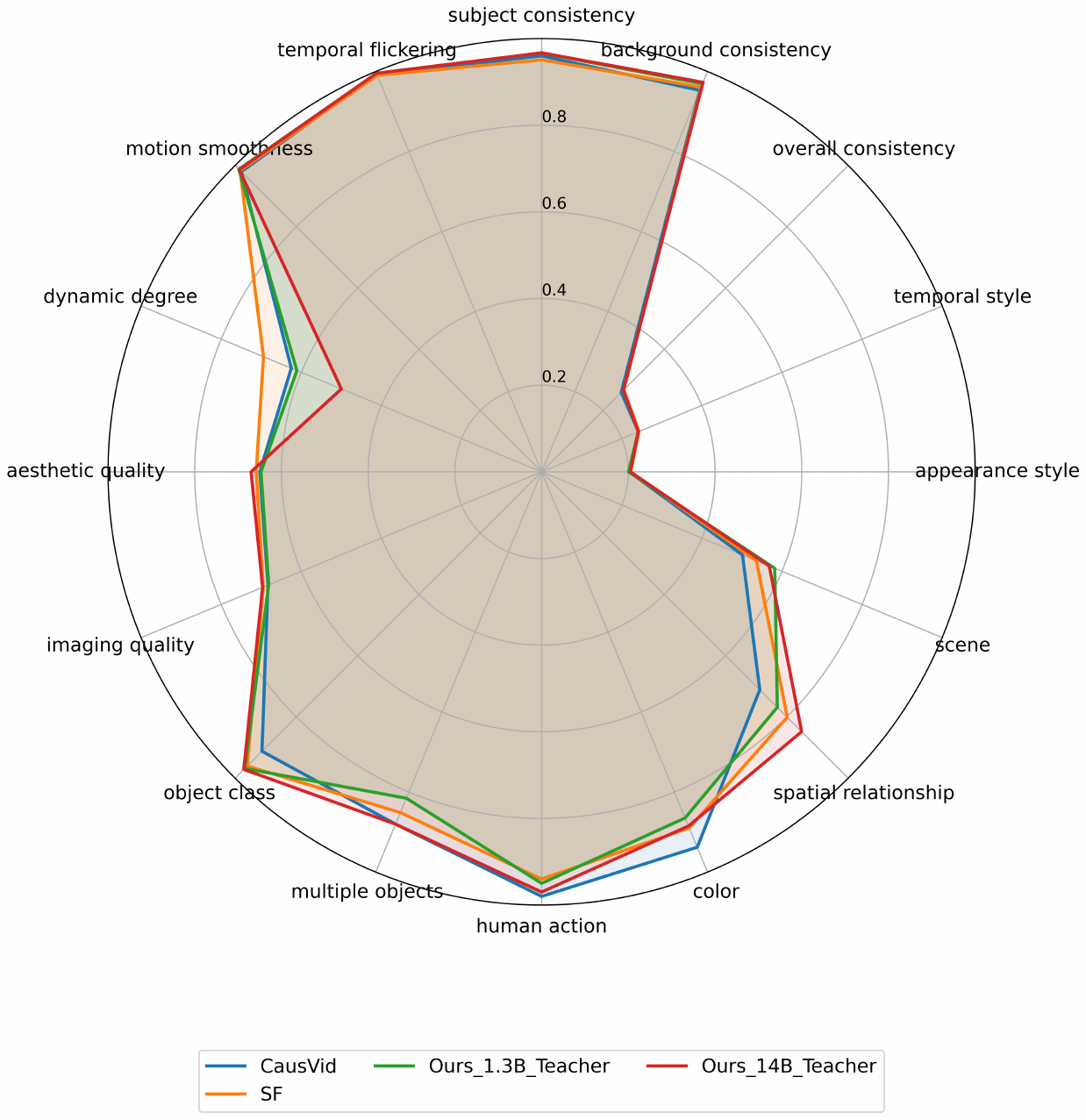}
    \caption{\textbf{VBench scores comparison} We compare \mname with CausVid~\cite{yin2025slow} and Self-Forcing~\cite{huang2025self} in 16 VBench scores.}
    \label{fig:vbench}
\end{figure}

\begin{figure*}[htbp]
    \centering
    \includegraphics[width=\textwidth]{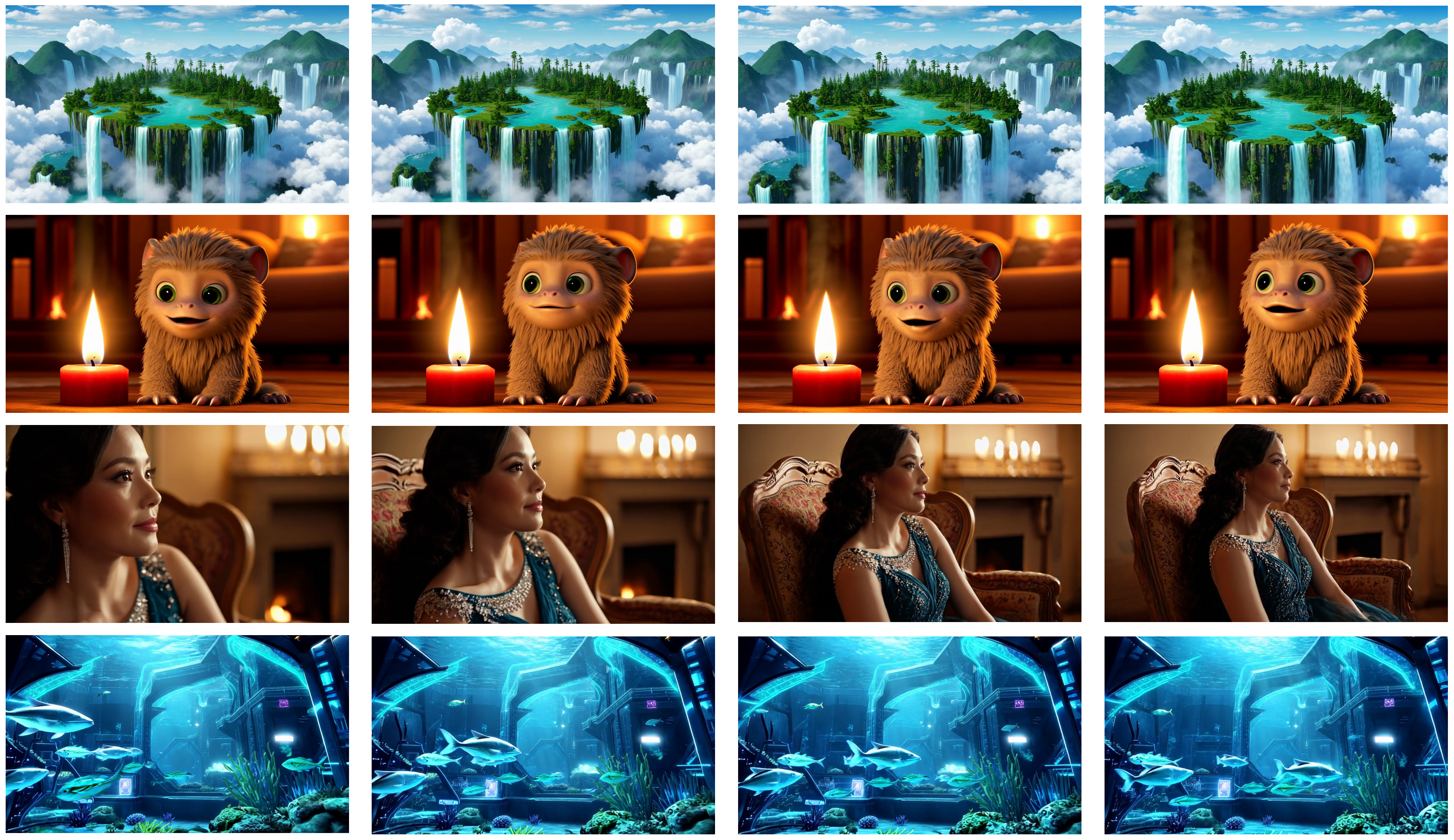}
    \caption{Additional qualitative results from \mname}
    \label{fig:qualitative-results-supp}
\end{figure*}

We report all 16 VBench metrics and compare them against CausVid~\cite{yin2025slow} and Self-Forcing~\cite{huang2025self}. As shown in Figure~\ref{fig:vbench}, \mname matches or surpasses existing models in both visual quality and temporal consistency across the full set of metrics, while maintaining efficient generation speed. Additional qualitative results are provided in Figure~\ref{fig:qualitative-results-supp}, demonstrating the high visual fidelity of our method. We also include video samples in the supplemental material.

\end{document}